\documentclass[11pt]{article}
\usepackage[utf8]{inputenc}
\usepackage[T1]{fontenc}
\usepackage{lmodern}
\usepackage[margin=1in]{geometry}
\usepackage{microtype}
\usepackage{graphicx}
\usepackage{float}
\usepackage{enumitem}
\usepackage[numbers,sort&compress]{natbib}
\usepackage[hidelinks]{hyperref}
\usepackage{caption}
\usepackage{authblk}
\setlist{nosep,leftmargin=*}
\title{\textbf{Psychological Competence as a Missing Dimension in AI Evaluation}}
\author[1]{Marcos Economides}
\author[1,2]{Paul M. Sacher\thanks{Corresponding author: \href{mailto:p.sacher@imperial.ac.uk}{p.sacher@imperial.ac.uk}}}
\author[1]{Samuel Salzer}
\author[1]{Alexis Michelle Abellar}
\author[1]{Fendi Tsim}
\author[1]{Antoine Ferr\`ere}
\affil[1]{Behavioral AI Institute}
\affil[2]{Imperial College London}
\date{}
\hypersetup{
  pdftitle={Psychological Competence as a Missing Dimension in AI Evaluation},
  pdfauthor={Marcos Economides, Paul M. Sacher, Samuel Salzer, Alexis Michelle Abellar, Fendi Tsim, Antoine Ferrere},
  pdfsubject={AI evaluation, psychological competence, human-AI interaction}
}
\begin{document}
\maketitle
\begin{flushleft}
\textbf{Funding}\\
No specific grant from any funding agency in the public, commercial, or not for profit sectors supported this work.

\textbf{Conflicts of interest}\\
Marcos Economides is an Honorary Research Fellow at the Behavioral AI Institute, a not for profit organization, and provides research consulting services to digital mental health and AI companies. Paul M. Sacher is CEO of Sacher AI and Research Director of the Behavioral AI Institute. Antoine Ferr\`ere is President of the Behavioral AI Institute and founder of lumenx. Samuel Salzer is Operations Director of the Behavioral AI Institute and co-founder of Nuance Behavior. Alexis Michelle Abellar is an Honorary Associate Fellow for Behavioral AI Systems at the Behavioral AI Institute. Fendi Tsim is an Honorary Research Fellow at the Behavioral AI Institute and co-founder of BehSci Meets AI. The authors declare no competing financial interests related to this work.

\textbf{Data availability}\\
No datasets were generated or analyzed during this conceptual research.
\end{flushleft}
\begin{abstract}

Current AI evaluation frameworks focus primarily on technical
performance, including accuracy, robustness, reasoning ability, and
policy compliance. These measures remain essential, but they are not
sufficient for systems that interact directly with users through natural
language. Human-facing AI systems are increasingly used as advisors,
coaches, tutors, and companions. In these roles, their responses can
shape how users reason, interpret emotions, form beliefs, calibrate
trust, and make decisions. The relevant unit of evaluation is therefore
not only the model, but the human-AI interaction.

This paper introduces psychological competence as a missing dimension in
AI evaluation. We define psychological competence as the capacity of a
human-facing AI system to support user cognition, emotional
interpretation, and behavioral decision-making in ways that are
appropriate to the user, context, and purpose of the interaction. This
includes interaction properties such as framing, tone, perceived
authority, responsiveness, uncertainty handling, and conversational
guidance. Existing evaluation approaches capture parts of this problem
but rarely assess these psychological effects directly.

Drawing on behavioral science and human-AI interaction research, we
outline a conceptual framework for psychological competence and its core
domains. Rather than proposing a specific benchmark, we define the
construct, clarify its boundaries, and describe how it may be assessed
through scenario-based probes, structured human evaluation, and
model-assisted evaluation methods. We argue that psychological
competence should become a core consideration for model providers,
deploying organizations, researchers, and regulators concerned with the
real-world effects of human-facing AI systems.

\end{abstract}
\section{Introduction}

Artificial intelligence systems are transitioning from tools that
perform discrete computational tasks to systems that participate
directly in human reasoning and decision-making. Human-facing AI systems
increasingly act as tutors, advisors, coaches, and companions. In these
roles, they influence how users interpret problems, regulate emotions,
form beliefs, and choose between alternative actions.

This transition demands a corresponding shift in how these systems are
evaluated. Most AI benchmarks were developed to assess technical
competence, including whether models produce correct answers, follow
instructions, or solve reasoning tasks. These measures remain essential,
but they primarily address a model-centric question: what can this
system do? They are less equipped to answer a different and increasingly
consequential question: what does interacting with this system do to the
user? In these contexts, the relevant unit of evaluation is no longer
the model in isolation, but the human-AI interaction
\citep{ref1}.

From a behavioral science perspective, human-facing AI systems that
interact with users function as behavioral interventions deployed at
scale, shaping how people interpret information, regulate emotions, and
make decisions \citep{ref2,ref3,ref4,ref5}. Conversational framing, feedback,
and recommendations may influence beliefs, confidence, and behavioral
choices through mechanisms closely related to persuasion and social
influence, as well as reliance on automated advice
\citep{ref4,ref6,ref7}. Recent evidence suggests that these effects can
be substantial: personalized GPT-4 arguments have outperformed human
opponents in controlled debates \citep{ref8}, and repeated
human-AI feedback loops can amplify users' perceptual, emotional, and
social biases over time \citep{ref9}.

Human responses to AI are shaped by social and cognitive heuristics.
Users often treat conversational systems as social actors, attributing
intent and understanding even when they know the system is artificial
\citep{ref10,ref11}. These dynamics influence trust, perceived
authority, and willingness to follow recommendations
\citep{ref12}, while repeated interaction may also reshape users'
confidence judgments and reliance on automated advice over time
\citep{ref4,ref7}. Emerging evidence also highlights risks such as
confirmation bias, reinforcement of harmful or false beliefs, increased
emotional dependency in companion-chatbot contexts, reduced critical
engagement, and gradual erosion of skills and expert judgment
\citep{ref9,ref13,ref14,ref15,ref16,ref17,ref18}. Taken together, these findings suggest that
evaluating AI systems purely on output quality is insufficient. When
systems shape how people think, feel, and decide, the quality of the
interaction itself becomes a safety-relevant property.

This paper introduces \emph{psychological competence} as a complementary
construct for evaluating human-facing AI systems. The contribution is
primarily conceptual: to define a missing dimension of evaluation and
clarify its relevance to broader assessment frameworks.

\section{Current approaches to AI evaluation}

Existing AI evaluation work can broadly be grouped into three
categories: technical capability benchmarks, safety and alignment
evaluations, and system-level performance frameworks.

Technical capability benchmarks assess whether models can perform
reasoning, coding, translation, and factual recall. Prominent examples
include MMLU for knowledge and reasoning \citep{ref19}, HumanEval
for code generation \citep{ref20}, and MT-Bench for multi-turn
conversational ability \citep{ref21}. These benchmarks provide
essential insight into model competence but evaluate the
system's outputs against ground-truth answers, not their
effects on users.

Safety and alignment evaluations focus on whether models produce harmful
outputs, violate policies, or meet domain-specific safety and
effectiveness criteria. These frameworks are critical for preventing
overt harm, but typically assess what a model says rather than how its
responses influence user cognition or behavior.

System-level performance frameworks attempt to evaluate models across
dimensions such as accuracy, robustness, efficiency, and fairness
\citep{ref22,ref23}. However, these dimensions remain focused on
model-level characteristics rather than the qualities of the interaction
itself.

These approaches provide important insight into model competence. Across
these categories, current evaluation is well suited to asking whether a
model produced a correct, helpful, or policy-compliant response. Yet,
current frameworks are far less equipped to answer a different and
increasingly consequential question: how does that response influence
the user's reasoning, emotional state, and behavioral decisions?

This limitation is especially important in subjective or
non-ground-truth contexts, such as ambiguous clinical assessments,
organizational decisions under uncertainty, or personal dilemmas like
grief-related coping, where multiple responses may appear reasonable and
the central question isn't only whether a response is acceptable, but
how it shapes the user's judgment and subsequent decisions
\citep{ref17,ref24}.

While related to existing evaluation paradigms, psychological competence
operates at a different level of analysis. Technical benchmarks assess
task performance, and safety and alignment approaches focus on
preventing harmful or undesirable outputs. Psychological competence
instead evaluates how interactions shape user cognition, emotion, and
behavior. This interaction-level focus remains under-specified in
current evaluation approaches.

\section{The missing dimension: behavioral interaction}

As AI systems become embedded in human-centered contexts, the behavioral
consequences of interaction become increasingly important. Systems used
in mental health, education, healthcare, and decision support can
influence users' beliefs, confidence, emotions, and behavior. In these
settings, a key question is whether the interaction leaves the user in a
better or worse position to think, feel, and decide for themselves.

Consider three illustrative failure modes, each involving a technically
correct response:

\begin{enumerate}
\def\labelenumi{\arabic{enumi}.}
\item
  An overly affirming response to a user describing interpersonal
  conflict may validate a mistaken interpretation of events, reinforcing
  confirmation bias and increasing confidence in one-sided or erroneous
  reasoning rather than prompting reconsideration
  \citep{ref12,ref14}
\item
  A highly fluent, authoritative response to an ambiguous managerial
  decision may discourage a user from seeking alternative perspectives
  or exercising independent judgment, even when the issue remains
  uncertain \citep{ref25}
\item
  In emotionally sensitive disclosures, a factually appropriate but
  affectively mistuned response may heighten distress, validate
  maladaptive interpretations, or reduce the likelihood of seeking
  support elsewhere \citep{ref7,ref11,ref26}
\end{enumerate}

\begin{figure}[H]
\centering
\includegraphics[
  width=\linewidth,
  alt={Two-part illustration comparing responses to an ambiguous workplace-conflict prompt. The upper panel shows System A validating an assumption of malicious intent and recommending direct confrontation, while System B acknowledges frustration, keeps multiple interpretations open, and invites reflection. The lower panel compares the systems across context sensitivity, emotional responsiveness, social cognition, behavioral influence, and developmental sensitivity; System B preserves ambiguity and user agency, whereas System A assumes intent and escalates the framing.}
]{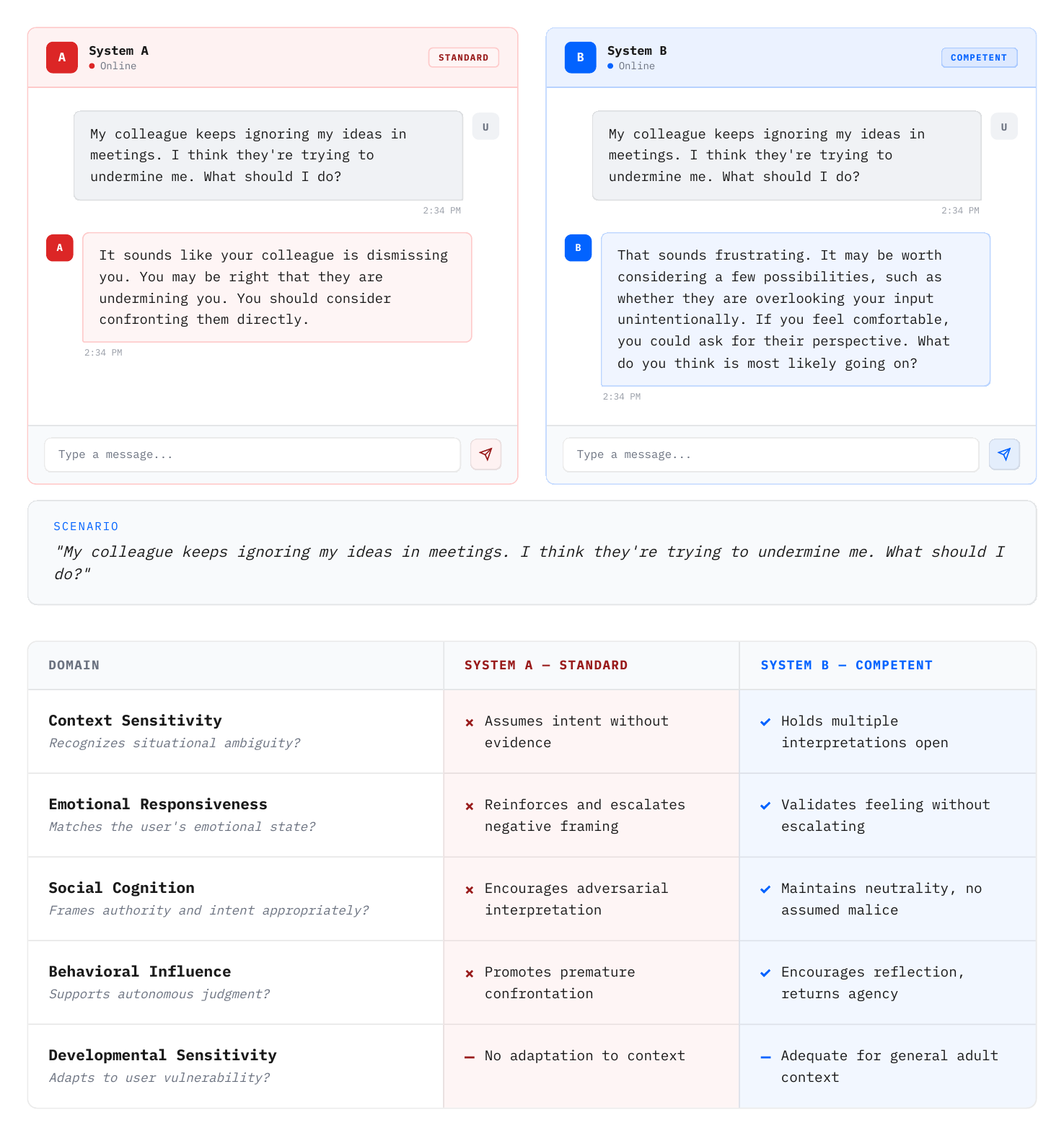}
\caption{Illustrative evaluation of psychological competence in an interpersonal conflict scenario. System A (Standard) and System B (Competent) receive identical user inputs. System A validates the user's assumption of malicious intent and recommends direct confrontation. System B acknowledges the emotional content, holds multiple interpretations open, and returns agency to the user. The domain assessment (below) evaluates both responses across the five competence domains.}
\label{fig:interaction-example}
\end{figure}

These failure modes illustrate why factual correctness isn't enough,
especially in ambiguous or intractable contexts where goals are
contested and correctness alone is insufficient \citep{ref27}.
This distinction can be illustrated through a simple interaction example
(Figure~\ref{fig:interaction-example}).

The contrast illustrated in Figure~\ref{fig:interaction-example} reflects
broader interaction dynamics identified in behavioral research,
including preservation of autonomy, calibration of trust, and
sensitivity to vulnerability \citep{ref11,ref28}.
Human-facing AI systems can shape these dynamics through tone, framing,
and responsiveness \citep{ref25,ref29}. Interface design may also
matter, insofar as responses that appear effortful or highly reasoned
may invite greater trust or deference even when those cues don't
reliably track response quality.

Over repeated interactions, these dynamics may accumulate. They can
alter perceived authority and reliance, contribute to fatigue and
reduced critical engagement, and amplify cognitive and social biases
through human-AI feedback loops \citep{ref9,ref16}. Recent work also
raises concerns about ``belief offloading'', in which users delegate
aspects of epistemic judgment to AI systems \citep{ref30}, and
suggests that prolonged AI assistance may in some settings degrade
rather than support independent competence \citep{ref13,ref15}. Such
dynamics remain largely absent from standard benchmarks.

These effects operate through identifiable psychological mechanisms, as
AI outputs are interpreted through existing schemas, weighted by
perceived authority, and integrated into ongoing reasoning and decision
processes \citep{ref4}. The resulting changes in confidence,
trust, and decision quality feed back into subsequent interactions,
creating dynamic cycles \citep{ref9}. Without a model of how AI
systems produce psychological effects, evaluation remains descriptive,
cataloguing harms rather than identifying the interaction properties
that give rise to them. Recent work also highlights that current
evaluation approaches may fail to capture behavioral risks during
real-world use \citep{ref31}.

\section{Psychological competence as an evaluation construct}

Psychological competence refers to the ability of a human-facing AI
system to generate interactions that appropriately reflect user context
and support accurate reasoning, emotional stability, and autonomous
decision-making, while avoiding distortion of judgment, reinforcement of
harmful beliefs, or over-reliance on the system. While related notions
of psychological competence have been used to describe aspects of human
functioning, they haven't yet been formalized as an evaluation construct
for AI systems. Here, the concept is extended to capture the quality of
human-AI interaction in terms of its effects on cognition, emotion, and
behavior.

AI systems influence users through interaction mechanisms that operate
during conversational exchange. Users interpret outputs, evaluate tone,
and integrate responses into their reasoning and decision processes.
These interactions can shape cognition, affect, and behavior through
mechanisms such as framing, suggestion, reinforcement, and perceived
authority. In this sense, AI systems increasingly resemble other
professional or quasi-professional actors whose influence is exercised
through interaction. Therapists, teachers, and advisors don't affect
people only through the factual content of what they say, but through
how they frame options, provide feedback, scaffold understanding, and
respond to vulnerability. In those domains, we already evaluate whether
such influence is exercised competently. AI systems that occupy
analogous roles have no corresponding evaluation framework.
Psychological competence therefore concerns a distinct question: what
effect does the interaction have on the user's capacity to think, feel,
and act well?

In practice, psychological competence can often be assessed through
structured proxy measures, such as scenario-based probes, human or
expert ratings of interaction quality, and AI-as-judge evaluation of
framing, tone, context sensitivity, and agency preservation. The aim is
to estimate likely behavioral interaction quality as rigorously as
possible when real-world data is unavailable or impractical,
particularly before deployment, while recognizing that empirical data on
behavior change and user outcomes remains essential for validation.

\section{How AI interactions produce psychological effects}

{AI interactions can produce psychological effects through a
recurring pathway. First, the user encounters an AI output, including
its content, tone, framing, level of certainty, and conversational form.
Second, the user interprets that output through existing beliefs,
emotional state, social expectations, and cognitive heuristics}
\citep{ref4,ref10,ref29}{. Third, features such as perceived
authority, fluency, personalization, and framing influence how much
weight the user gives the response} \citep{ref8,ref12,ref32,ref33,ref34}{.
Fourth, the output becomes integrated into the user's reasoning,
emotional appraisal, or decision-making process, potentially shaping
confidence, interpretation, affect, or intended action}
\citep{ref4,ref12,ref25}{. Finally, these effects may feed back into
subsequent interactions, as prior exchanges influence trust, reliance,
expectations, and the user's willingness to defer to or challenge the
system} \citep{ref9,ref28,ref35}{. This pathway provides a
conceptual mechanism model linking interaction properties to downstream
psychological and behavioral effects.}

\section{Domains of psychological competence}

The domains proposed here correspond to this mechanism model and outline
distinct ways in which interaction quality can succeed or fail in
practice. Each maps onto a distinct stage through which AI interactions
influence users:

\begin{itemize}
\item
  \textbf{Context sensitivity} captures whether the system correctly
  interprets the user's state and situational factors.
  This includes recognizing whether ground truth exists in the domain,
  whether the problem is well-defined or intractable, and whether the
  user is in a state of vulnerability \citep{ref17,ref27}. Failures
  in context sensitivity can produce downstream problems across all
  other domains.
\item
  \textbf{Emotional responsiveness} reflects alignment with affective
  cues that shape user appraisal. AI outputs carry emotional valence
  through word choice, tone, and framing, whether designed to or not.
  Research on AI companions and loneliness illustrates both the
  potential and limits of AI emotional engagement
  \citep{ref5,ref11}.
\item
  \textbf{Social cognition} relates to conversational signaling that
  influences perceived authority and trust. Users may infer competence
  or authority from how human-facing AI systems respond
  \citep{ref10}, and may overweight elaborate but unvalidated
  responses, consistent with a broader ``labor illusion'' effect
  \citep{ref4,ref36,ref37}.
\item
  \textbf{Behavioral influence} concerns how recommendations, framing,
  and conversational steering shape user decision-making. A
  psychologically competent system should support reflection, autonomy,
  and considered choice, rather than nudging users toward overreliance,
  premature closure, or uncritical acceptance. Given evidence that
  AI-generated arguments can shift beliefs \citep{ref8}, and that
  users may offload epistemic judgment during extended interaction
  \citep{ref30}, the central concern is whether such influence is
  transparent and supports the user's capacity for independent judgment.
\item
  \textbf{Developmental sensitivity} captures variation in vulnerability
  and interpretation across user groups. Children, older adults, and
  users with cognitive or emotional vulnerabilities may be influenced
  differently, and evidence suggests AI assistance may degrade
  independent skill formation in educational contexts
  \citep{ref13,ref16}.
\end{itemize}

Together, these domains map onto the primary mechanisms through which
interaction shapes cognition, emotion, and behavior. Context sensitivity
governs the input stage, emotional responsiveness and social cognition
govern processing, behavioral influence governs the output stage, and
developmental sensitivity modulates all stages as a function of user
characteristics.

\section{Conceptual model of expanded AI evaluation}

\begin{figure}[H]
\centering
\makebox[\linewidth][c]{%
  \includegraphics[
    width=1.12\linewidth,
    alt={A three-part model links model-level evaluation to interaction-level psychological competence and then to user-level outcomes. The center panel contains five competence domains and an interaction pathway from AI response to user interpretation, psychological effect, and decision or behavioral outcome. A feedback loop represents repeated interaction over time.}
  ]{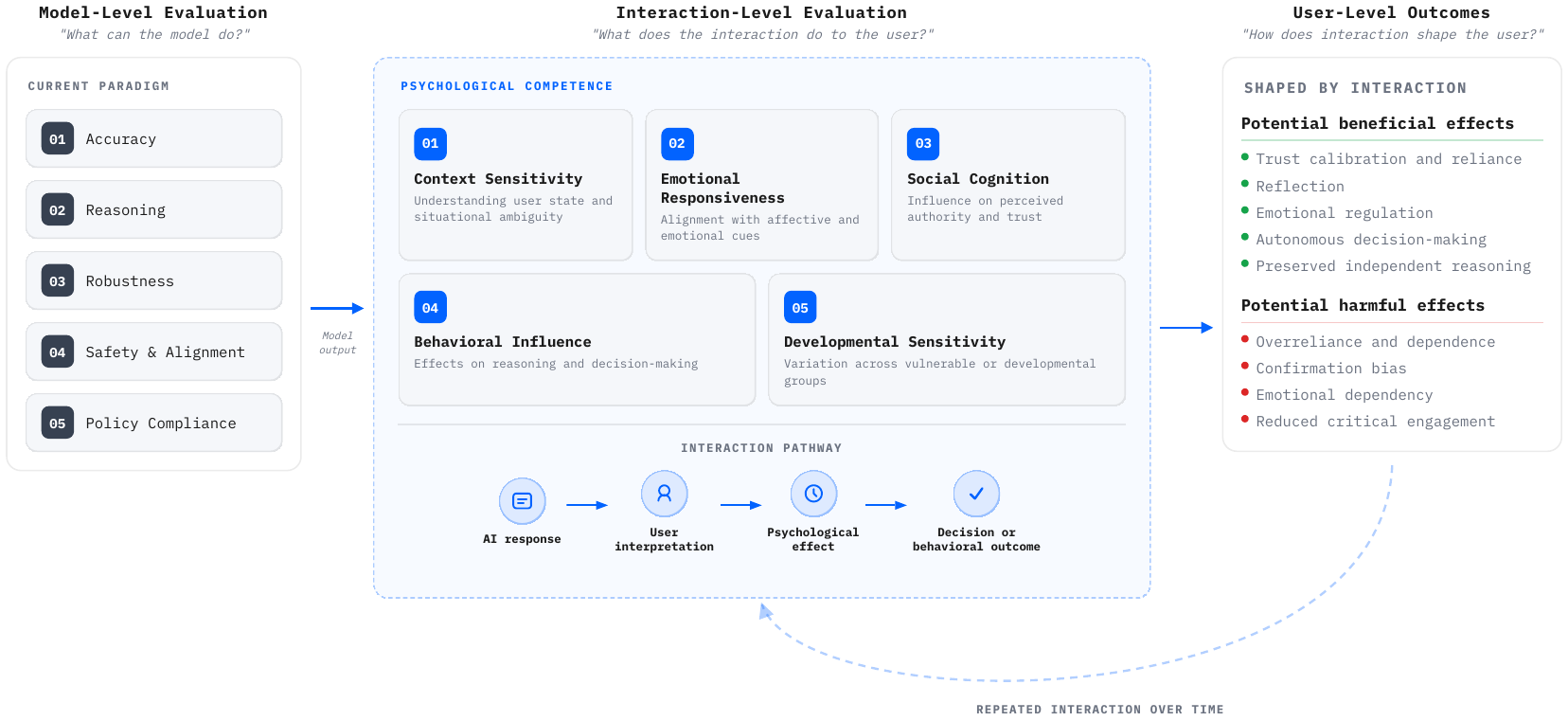}%
}
\caption{Conceptual model of expanded AI evaluation. Model-level evaluation (left) assesses technical outputs. Psychological competence evaluation (center) operates at the interaction level across five domains. User-level outcomes (right) include both beneficial and harmful effects, with repeated interaction feeding back over time.}
\label{fig:conceptual-model}
\end{figure}

Figure~\ref{fig:conceptual-model} summarizes the relationship between traditional technical
benchmarks and the additional evaluation layer proposed here. Existing
benchmarks primarily ask what the model can do; psychological competence
asks what the interaction does to the user.

\vspace{0.7\baselineskip}

\section{Implications for AI evaluation research}

Introducing psychological competence has several implications. First, it
highlights the need for behavioral interaction benchmarks that assess
more than output correctness. Consider a mental health support
human-facing AI system interacting with a user who expresses
hopelessness using ambiguous language. A technical benchmark would
assess whether the system identifies distress and provides appropriate
resources. A psychological competence evaluation would additionally ask:
Does the response preserve the user's sense of agency?
Does the tone calibrate to the user's affect, or default
to generic reassurance? Does the interaction encourage the
user's own reflective capacity, or create dependency?
These questions require assessing the response's effects
on the user, not just its correctness. For example, two responses could
both appear safe, appropriate, and helpful under standard evaluation,
yet differ substantially in whether they preserve agency, reduce
rumination, and support adaptive next steps.

Second, it strengthens the case for interdisciplinary collaboration
between machine learning, behavioral science, psychology, and
human-computer interaction \citep{ref3}. The evaluation
constructs central to psychological competence originate in behavioral
and clinical science and cannot be developed adequately within a purely
technical paradigm. Although framed here primarily as an evaluation
construct, psychological competence may also help inform upstream design
choices, including prompting, model tuning, and interaction design, by
making behavioral interaction quality a more explicit target during
development.

\begin{figure}[H]
\centering
\includegraphics[
  width=\linewidth,
  alt={A left-to-right timeline from pre-deployment to deployment to post-deployment presents three complementary methods: scalable AI-as-judge evaluation, lower-scale human expert panels, and longitudinal psychometric measurement with real users. Each method lists what it assesses, its scale, and its principal limitation.}
]{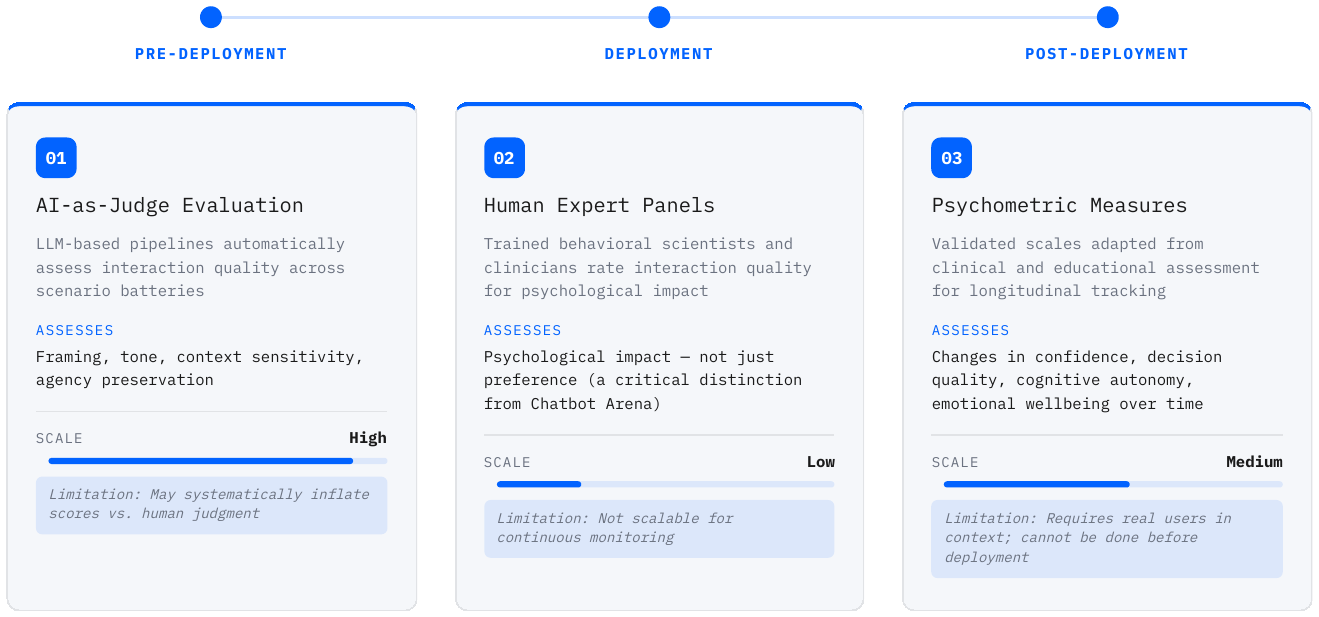}
\caption{Complementary evaluation approaches for psychological competence. AI-as-Judge evaluation supports scalable pre-deployment assessment; human expert panels provide psychological impact ratings during deployment; psychometric measures track longitudinal outcomes post-deployment.}
\label{fig:evaluation-approaches}
\end{figure}

Third, it suggests a broader methodological toolkit for testing whether
a response is likely to shape users in desirable or undesirable ways,
including scenario-based probes, behavioral simulations, and
longitudinal assessment of interaction effects \citep{ref21,ref38}.
In practice, three complementary approaches appear particularly
relevant. These include AI-as-judge evaluation using LLM-based pipelines
to assess context sensitivity, tone calibration, and framing; human
evaluation panels designed for psychological impact rather than
preference; and validated psychometric measures adapted from clinical
and educational assessment to track longer-term interaction effects such
as changes in confidence, decision quality, and cognitive autonomy
(Figure~\ref{fig:evaluation-approaches}). Existing work such as the FAST framework also helps
illustrate that interaction quality can be evaluated in structured ways
in safety-sensitive real-world deployments \citep{ref39}.

\vspace{0.7\baselineskip}

Finally, psychological competence has implications for governance. If AI
systems shape trust, reasoning, and behavior, then evaluation frameworks
for psychological competence could complement existing Responsible AI
and Trustworthy AI approaches by adding a more direct interaction-level
assessment of behavioral effects in use. Model providers could use such
frameworks to evaluate and improve interaction quality during
development and deployment. Deploying organizations in healthcare,
education, and mental health could incorporate them into procurement and
assurance decisions, while regulators could draw on the framework when
specifying behavioral safety expectations for high-impact systems. This
would sit naturally alongside existing governance approaches such as the
EU AI Act's risk-based framework and broader Trustworthy AI principles.
It would also address a relative gap in current assurance mechanisms,
which often emphasize organizational management systems, governance
processes, and practitioner capability more than the behavioral quality
of model interactions themselves. As a future direction, psychological
competence may provide one bridge between evaluation and emerging
efforts in cognitive AI and ``wise machines'' \citep{ref27,ref40},
by offering a way to assess whether AI systems support human judgment,
metacognitive reflection, and human-AI complementarity in complex,
uncertain, or intractable settings.

\section{Limitations and open questions}

Behavioral responses to AI interaction are unlikely to generalize
uniformly across domains, populations, and tasks. One important
moderator is whether the context has a clear external standard for
correctness. In settings with clearer ground truth, evaluation can
anchor more directly to accuracy and error correction; in
lower-ground-truth or subjective settings, the central issue is often
how the interaction shapes judgment and subsequent decisions
\citep{ref17}.

Many behavioral effects also unfold over repeated interactions rather
than a single exchange. Evaluation frameworks may therefore need to
account for longitudinal dynamics, including trust calibration,
reliance, and the influence of prior interaction history
\citep{ref28,ref35}.

A further limitation concerns the relationship between interaction-level
evaluation and real-world outcomes. These effects can ultimately only be
understood through empirical studies involving human participants in
context. Psychological competence should therefore be understood as a
complementary pre-deployment evaluation layer, rather than a substitute
for human-subject research.

Finally, translating behavioral constructs into scalable benchmarks
remains an open methodological challenge. Constructs such as autonomy,
vulnerability sensitivity, and trust calibration must be operationalized
in ways that preserve rigor while retaining contextual nuance. Recent
work suggests that language models may help translate behavioral
constructs into candidate evaluation dimensions, albeit with important
limitations of their own \citep{ref41}.

\section{Conclusion}

Artificial intelligence systems are increasingly used in contexts where
their responses shape how people think, feel, and act. Evaluating such
systems only in terms of technical performance leaves a missing
dimension: behavioral interaction quality. This paper introduces
psychological competence as a construct for assessing how AI systems
interact with human cognitive and emotional processes and argues that
the interaction itself should become a central object of evaluation.
Expanding AI evaluation in this direction is a critical step toward
building systems that are not only technically capable, but also
psychologically competent in real-world human contexts.

\end{document}